\documentclass[sn-mathphys]{sn-jnl}
\usepackage{hyperref}
\usepackage{url}
\usepackage{makeidx}
\usepackage{booktabs}
\usepackage{enumerate}
\usepackage{stfloats}
\usepackage{graphicx}
\usepackage{subfigure} 
\usepackage{float}
\usepackage{amssymb}
\usepackage[misc]{ifsym}
\usepackage{graphicx}
\usepackage{pifont}
\graphicspath{{image/}}
\usepackage{algorithm}  
\usepackage{algorithmicx}  
\usepackage{algpseudocode}  
\usepackage{amsmath}  
\floatname{algorithm}{Algorithm}

\jyear{2021}

\begin{document}
\title[Autonomous Intelligent Systems]{Multi-agent Reinforcement Learning for Cooperative Lane Changing of Connected and Autonomous Vehicles in Mixed Traffic}

\author[1]{\fnm{Wei} \sur{Zhou}}\email{zhouwei0330@tongji.edu.cn}
\equalcont{Both authors contributed equally to this work.}
\author[2]{\fnm{Dong} \sur{Chen}}\email{chendon9@msu.edu}
\equalcont{Both authors contributed equally to this work.}
\author*[1]{\fnm{Jun} \sur{Yan}}\email{yanjun@tongji.edu.cn}
\author[2]{\fnm{Zhaojian} \sur{Li}}\email{lizhaoj1@msu.edu}
\author[1]{\fnm{Huilin} \sur{Yin}}\email{yinhuilin@tongji.edu.cn}
\author[1]{\fnm{Wanchen} \sur{Ge}}\email{gwc828@tongji.edu.cn}

\affil[1] {\orgdiv{School of Electronic and Information Engineering}, \orgname{Tongji University}, \orgaddress{\street{Caoangong Street}, \city{Shanghai}, \postcode{201804},  \country{China}}}

\affil[2]{\orgdiv{Mechanical Engineering}, \orgname{Michigan State University}, \orgaddress{\city{Lansing}, \postcode{48824},  \country{USA}}}

\abstract{
Autonomous driving has attracted significant research interests in the past two decades as it offers many potential benefits, including releasing drivers from exhausting driving and mitigating traffic congestion, among others. Despite promising progress, lane-changing remains a great challenge for autonomous vehicles (AV), especially in mixed and dynamic traffic scenarios. Recently, reinforcement learning (RL) has been widely explored for lane-changing decision makings in AVs with encouraging results demonstrated. However, the majority of those studies are focused on a single-vehicle setting, and lane-changing in the context of multiple AVs coexisting with human-driven vehicles (HDVs) have received scarce attention. In this paper, we formulate the lane-changing decision-making of multiple AVs in a mixed-traffic highway environment as a multi-agent reinforcement learning (MARL) problem, where each AV makes lane-changing decisions based on the motions of both neighboring AVs and HDVs. Specifically, a multi-agent advantage actor-critic (MA2C) method is proposed with} a novel local reward design and a parameter sharing scheme. In particular, a multi-objective reward function is designed to incorporate fuel efficiency, driving comfort, and the safety of autonomous driving. A comprehensive experimental study is made that our proposed MARL framework consistently outperforms several state-of-the-art benchmarks in terms of efficiency, safety, and driver comfort.


\maketitle
\section{Introduction}
\label{sec1}
Autonomous driving has received significant research interest in the past two decades due to its many potential societal and economical benefits. Compared to traditional vehicles, autonomous vehicles (AVs) not only promise fewer emissions \cite{Paden} but are also expected to improve safety and efficiency. However, there exists a huge challenge in the task of high-level decision-making in AVs due to the complex and dynamic traffic environment, especially in mixed traffic co-existing with other road users. Lane changing is one of the largest challenges in the high-level decision-making of AVs, which has significant influences on traffic safety and efficiency \cite{desiraju2014minimizing, li2020cooperative}.

\begin{figure}[!ht]
  \centering
  \includegraphics[ width=0.7\textwidth]{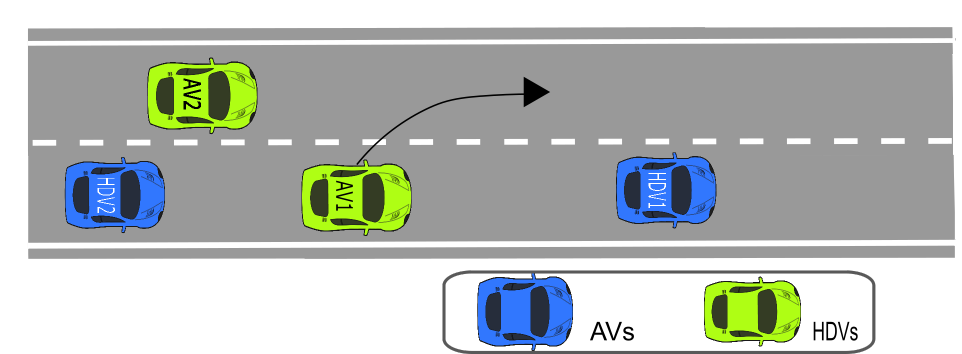}
  \caption{Illustration of the considered lane-changing scenario (green: AVs, blue: HDVs, arrow curve: a possible trajectory of the ego vehicle AV1 to make the lane change).}
  \label{lane-changing scene}
\end{figure}

The considered lane-changing scenario is illustrated in Fig.~\ref{lane-changing scene}, where AVs and HDVs co-exist on a one-way highway with two lanes. The AVs aim to safely travel through the traffic while making necessary lane changes to overtake slow-moving vehicles for improved efficiency. Furthermore, in the presence of multiple AVs, the AVs are expected to collaboratively learn a policy to adapt to HDVs and enable safe and efficient lane changes. As HDVs bring unknown/uncertain behaviors, planning, and control in such mixed traffic to realize safe and efficient maneuvers is a challenging task \cite{chen2020autonomous}. 


Recently, reinforcement learning (RL) has emerged as a promising framework for autonomous driving due to its online adaptation capabilities and
the ability to solve complex problems \cite{wang2019continuous, xi2020efficient}. 
Several recent studies have explored the use of RL in AV lane-changing \cite{wang2021harmonious, du2020cooperative, chen2020autonomous}, which consider a single AV setting where the ego vehicle learns a lane-changing behavior by taking all other vehicles as part of the driving environment for decision making. While this single-agent approach is completely scalable, it will lead to unsatisfactory performance in the complex environment like multi-AV lane-changing in mixed traffic that requires close collaboration and coordination among AVs \cite{Hoel}.  

On the other hand, multi-agent reinforcement learning (MARL) has been greatly advanced and successfully applied to a variety of complex multi-agent systems such as games \cite{vinyals2019grandmaster}, traffic light control \cite{Chu} and fleet management \cite{lin2018efficient}. The MARL algorithms have also been applied to autonomous driving \cite{shalev2016safe, wangmulti, ha2020leveraging, palanisamy2020multi}, with the objective of accomplishing autonomous driving tasks cooperatively and reacting timely to HDVs.
\textcolor{blue}{In particular, the MARL methods \cite{chen2021graph, ha2020leveraging} have been applied} to highway lane change tasks and show promising and scalable performance, in which AVs learn cooperatively via sharing the same objective (i.e., reward/cost function) that considers safety and efficiency. However, those reward designs often ignore the passengers' comfort, which may lead to sudden acceleration and deceleration that can cause ride discomfort. In addition, they assume that the HDVs follow unchanged, universal human driver behaviors, which is clearly oversimplified and impractical in the real world as different human drivers tend to behave quite differently. Learning algorithms should thus work with different human driving behaviors, e.g., aggressive or mild behaviors.

To address the above issues, we develop a multi-agent reinforcement learning algorithm by employing a multi-agent advantage actor-critic network (MA2C) for multi-AV lane-changing decision making, featuring a novel local reward design that incorporates the safety, efficiency, and passenger comfort as well as a parameter sharing scheme to foster inter-agent collaborations. The main contributions and the technical advancements of this paper are summarized as follows.
\begin{enumerate}
\item We formulate the multi-AV highway lane changing in the mixed traffic as a decentralized cooperative MARL problem, where agents cooperatively learn a safe and efficient driving policy.
\item We develop a novel, efficient, and scalable multi-agent advantage actor-critic network model, by introducing a parameter-sharing mechanism and effective reward function design.
\item We conduct a comprehensive empirical study on three different traffic densities and two levels of drivers' behavior modes and compare with other state-of-the-art models to demonstrate the driving safety, efficiency, and driver comfort of our models.
\end{enumerate}

The rest of the paper is organized as follows. Section~\ref{section:RELATED WORK} reviews the state-of-the-art dynamics-based and RL/MARL algorithms for autonomous driving tasks. The preliminaries of RL and the proposed MARL algorithm are introduced in Section~\ref{section:PROPOSED SCHEMES}. Experiments, results, and discussions are presented in Section~\ref{section:NUMERICAL EXPERIMENTS AND DISCUSSION}. Finally, we summarize the paper and discuss future work in Section~\ref{section:Conclusion}. 

\section{Related Work}
\label{section:RELATED WORK}
In this section, we survey the existing literature on decision-making tasks in autonomous driving, which can be mainly classified into two categories: non-data-driven and data-driven methods.

\subsection{Non-data-driven Methods}
Conventional rule-based or model-based approaches \cite{ho2009lane, nilsson2016if, nilsson2016lane} rely on hard-coded rules or dynamical models to construct predefined logic mechanisms to determine the behaviors of ego vehicles under different situations.  For instance, lane changing guidance can be realized by establishing virtual trajectory references for every vehicle, and a safe trajectory is then planned by considering the trajectories of other vehicles \cite{ho2009lane}.In the previous literature \cite{nilsson2016if}, a low-complexity lane-changing algorithm was developed by following heuristic rules such as keeping appropriate inter-vehicle traffic gaps and time instances to perform the maneuver. After that, an optimization-based lane change approach was proposed \cite{nilsson2016lane}, which formulates the trajectory planning problem as coupled longitudinal and lateral predictive control problems and is then solved via Quadratic Programs under specific system constraints. However, the rules and optimization criteria for real-world driving problems may become too complex to be explicitly formulated for all scenarios. Such a problem is more serious in mixed-traffic scenarios with unknown or stochastic behaviors of drivers.

\subsection{Data-Driven Methods}
Recently, data-driven methods, such as reinforcement learning (RL), have received great attention and have been widely explored for autonomous driving tasks. Particularly, a model-free RL approach based on deep deterministic policy gradient (DDPG) was proposed in the recent literature \cite{wang2019continuous} to learn a continuous control policy efficient lane changing. Afterward, a safe RL framework \cite{chen2020autonomous} was presented by integrating a lane-changing regret model into a safety supervisor based on an extended double deep Q-network (DDQN). In the significant literature \cite{chen2019attention}, a hierarchical RL algorithm was developed to learn lane-changing behaviors in dense traffic by applying the designed temporal and spatial attention strategies, and promising performance is demonstrated in the TORCS simulator under various lane change scenarios. However, the aforementioned methods are designed for the single-agent (i.e., one ego vehicle) scenarios, treating all other vehicles as part of the environment, which makes them implausible for the considered multi-agent lane-changing setting where collaboration and coordination among AVs are required. 

On the other hand, multi-agent reinforcement learning (MARL) algorithms have also been explored for autonomous driving tasks \cite{shalev2016safe, dong2020drl, wangmulti, chen2021deep}. A MARL algorithm with hard-coded safety constraints \cite{shalev2016safe} was proposed to solve the double-merge problem. In such a framework, a hierarchical temporal abstraction method was applied to reduce the effective horizon and the variance of the gradient estimation error. In the recent literature \cite{chen2021deep}, a MARL algorithm was delivered to solve the on-ramp merging problem with safety enhancement by a novel priority-based safety supervisor. In addition, a novel MARL approach \cite{dong2020drl} was realized with the combination of Graphic Convolution Neural Network (GCN) \cite{kipf2016semi} and Deep Q Network (DQN) \cite{mnih2013playing} to better fuse the acquired information from collaborative sensing, showing promising results on a 3-lane freeway containing 2 off-ramps highway environment. While these MARL algorithms only consider the efficiency, and safety in their designed reward function, another important factor, the passenger comfort, is not considered in their reward function design. Furthermore, those approaches assume the HDVs follow a constant, universal driving behavior, which has limited implications for real-world applications as different human drivers may behave totally differently.

In this paper, we formulate the decision-making of multiple AVs on highway lane changing as a MARL problem, where a multi-objective reward function is proposed to simultaneously promote safety, efficiency, and passenger comfort. A parameter-sharing scheme is exploited to foster inter-agent collaborations. Experimental results on three different traffic densities with two levels of driver aggressiveness show the proposed MARL performs well on different lane change scenarios.

\section{Problem Formulation}
\label{section:PROPOSED SCHEMES}
In this section, we review the preliminaries of RL and formulate the considered highway lane-changing problem as a partially observable Markov decision process (POMDP). Then we present the proposed multi-agent actor-critic algorithm, featuring a parameter-sharing mechanism and efficient reward function design, to solve the formulated POMDP.

\subsection{Preliminary of RL}
In the standard RL setting, the agent aims to learn an optimal policy $\pi^*$ to maximize the accumulated future reward $R_t = \sum_{k=0}^{T} \gamma ^k r_{t+k}$ from the time step $t$ with discount factor $\gamma\in (0,1]$ by continuous interacting with the environment. Especially, at time step $t$, the agent receives a state $s_t \in \mathcal{R}^{n}$ from the environment and selects an action $a_t \in\mathcal{A}^{m}$ according to its policy $\pi:\,\mathcal{S}\rightarrow\Pr(\mathcal{A})$. As a result, the agent receives the next state $s_{t+1}$ and receives a scalar reward $r_t$. If the agent can only observe a part of the state $s_t$, the underlying dynamics becomes a POMDP \cite{spaan2012partially} and the goal is then to learn a policy that maps from the partial observation to an appropriate action to maximize the rewards. 

The action-value function $Q^{\pi}(s, a) = E[R_t{\mid}{s=s_t}, a]$ is defined as the expected return obtained by selecting an action $a$ in state $s_t$ and following policy $\pi$ afterwards. The optimal Q-function is given by $Q^{*}(s,a) = \max_{\pi} Q^{\pi}(s,a)$ for state $s$ and action $a$. Similarly, the state-value function is defined as $V^{\pi}(s_t) = E_{\pi}{[R_t{\mid}{s=s_t}]}$ representing the expected return for following the policy $\pi$ from state $s_t$. In model-free RL methods, the policy is often represented by a neural network denoted as $\pi_{\theta}(a_t{\mid}s_t)$, where $\theta$ is the learnable parameters. In actor-critic (A2C) algorithms \cite{mnih2016asynchronous}, a critic network, parameterized by $\omega$, learns the state-value function $V_{\omega}^{\pi_\theta}(s_t)$ and an actor network $\pi_{\theta}(a_t{\mid}s_t)$ parameterized by $\theta$ is applied to update the policy distribution in the direction suggested by the critic network as follows:
\begin{equation}\label{eqn:advantagepolicygradient}
\theta \leftarrow \theta + E_{\pi_{\theta}} \left[ \Big (\nabla_{\theta} \log \pi_{\theta}(a_t{\mid}s_t) \Big)
 A_t \right],
\end{equation}
where the advantage function $A_t= Q^{\pi_\theta}(s,a) - V_{\omega}^{\pi_\theta}(s_t)$ \cite{mnih2016asynchronous} is introduced to reduce the sample variance. 
The parameters of the state-value function are then updated by minimizing the following loss function: 
\begin{equation}\label{eqn:valueloss}
\min_{\omega} E_{\mathcal{B}}\Big (R_t + \gamma V_{\omega'} (s_{t+1}) - V_{\omega}(s_t)\Big )^2,
\end{equation} 
where $\mathcal{B}$ is the experience replay buffer that stores previously encountered trajectories and $\omega'$ denotes the parameters of the target network \cite{mnih2013playing}.

\subsection{Lane Changing as MARL}
In this subsection, we develop a decentralized, MARL-based approach for highway lane-changing of multiple AVs. Discontinuous evaluation is a very common way to design in the autonomous driving field, which is widely used by many papers \cite{DBLP:journals/corr/abs-1811-07214,Schester2019LongitudinalPC,Palanisamy2020MultiAgentCA,Wang2019ContinuousCF,Mavrogiannis2020BGAPBA}. In particular, we model the mixed-traffic lane-changing environment as a multi-agent network: $\mathcal{G} = (\text{$\nu$}, \text{$\varepsilon$})$, where each agent (i.e., ego vehicle) $i \in \text{$\nu$}$ communicates with its neighbors $\mathcal{N}_i$ via the communication link $\varepsilon_{ij}\in\large\text{$\varepsilon$}$. The corresponding POMDP is characterized as $(\{\mathcal{A}_i, \mathcal{O}_i, \mathcal{R}_i\}_{i\subseteq \nu}, \mathcal{T})$, where $\mathcal{O}_i \in \mathcal{S}_i$ is the partial description of the environment state as stated in \cite{chu2020multi}. In a multi-agent POMDP, each agent $i$ follows a decentralized policy $\pi_i: \mathcal{O}_i \times \mathcal{S}_i \rightarrow [0, 1]$ to choose the action $a_t$ at time step $t$. The described POMDP is defined as:
\begin{enumerate}
\item \textit{State Space}:
The state space $\mathcal{O}_{i}$ of Agent $i$ is defined as a matrix $\mathcal{N}_{N_{i}}\times \mathcal{F}$, where $\mathcal{N}_{N_{i}}$ is the number of detected vehicles, and $\mathcal{F}$ is the number of features, which is used to represent the current state of vehicles. It includes the longitudinal position $x$, the lateral position $y$ of the observed vehicle relative to the ego vehicle, the longitudinal speed $v_x$, and the lateral speed $v_y$ of the observed vehicle relative to the ego vehicle.

\item \textit{Action Space}:
The action space $\mathcal{A}_i$ of agent $i$ is defined as a set of high-level control decisions, including speed up, slow down, cruising, turn left, and turn right. The action space combination for AVs is defined as $\mathcal{A}=\mathcal{A}_{1}\times \mathcal{A}_{2}\times \cdot\cdot\cdot\times \mathcal{A}_{N}$, where $N$ is the total number of vehicles in the scene.

\item \textit{Reward Function}:
Multiple metrics including safety, traffic efficiency, and passenger's comfort are considered in the reward function design:
\begin{itemize}
\item[$\bullet$] safety evaluation $r_{s}$: The vehicle should operate without collisions.

\item[$\bullet$] headway evaluation $r_{d}$: The vehicle should maintain a safe distance from the preceding vehicles during driving to avoid collisions.

\item[$\bullet$] speed evaluation $r_{v}$: Under the premise of ensuring safety, the vehicle is expected to drive at a high and stable speed.

\item[$\bullet$] driving comfort $r_{c}$: Smooth acceleration and deceleration are expected to ensure safety and comfort. In addition, frequent lane changes should be avoided.
\end{itemize}
As such, a multi-objective reward $r_{i,t}$ at the time step $t$ is defined as:
\begin{equation}
\label{eqn:reward}
r_{i,t}=\omega_{s}r_{s}+\omega_{d}r_{d}+\omega_{v}r_{v}-\omega_{c}r_{c},
\end{equation}
where $\omega_{s}$, $\omega_{d}$, $\omega_{v}$ and $\omega_{c}$ are the weighting coefficients. We set the safety factor $\omega_{s}$ to a large value, because safety is the most important criterion during driving. The details of the four performance measurements are discussed next:

\begin{enumerate}[(1)]
\item If there is no collision, the collision evaluation $r_{s}$ is set to 0, otherwise, $r_{s}$ is set as -1.
\item The headway evaluation is defined as
\begin{equation}
\label{equation:rd}
    r_{d}=\log\frac{d_{headway}}{v_{t}t_{d}},
\end{equation}
where $d_{headway}$ is the distance to the preceding vehicle, and $v_t$ and $t_d$ are the current vehicle speed and time headway threshold, respectively.

\item The speed evaluation $r_{v}$ is defined as
\begin{equation}
    r_{v}=\min\left \{\frac{v_{t}-v_{min}}{v_{max}-v_{min}},1\right\},
\end{equation}
where $v_{t}$, $v_{min}$ and $v_{max}$ are the current, minimum, and maximum speeds of the ego vehicle, respectively. Within the specified speed range, higher speed is preferred to improve the driving efficiency.

\item The driving comfort $r_{c}$ is defined as
\begin{equation}
\label{eqn:comfort}
    r_{c}=r_{a}+r_{lc},
\end{equation}
where $$r_{a}=\left\{
\begin{array}{rcl}
-1,&\vert a_{t}\vert \geq a_{th}\\
0,&\vert a_{t}\vert < a_{th}
\end{array}
\right.$$ is the penalty term of rapid acceleration and deceleration than a given threshold $a_{th}$. Here $a_{t}$ presents the acceleration at time $t$. 

$$r_{lc}=\left\{
\begin{array}{rcl}
-1,&\rm\normalsize change&\rm\normalsize lane\\
0,&\rm\normalsize keep&\rm\normalsize lane
\end{array}
\right.$$ is defined as the lane change penalty. Excessive lane changes can cause discomfort and safety issues. Note that this lane-changing penalty term is to avoid frequent, unnecessary lane changes while necessary lane changes (i.e., to maintain safety and efficiency) are still promoted through the safety and speed evaluation terms.
\end{enumerate}

\item \textit{Transition Probability}: The transition probability $T(s^{,}\mid s,a)$ characterizes the transition from one state to another. Since our MARL algorithm is a model-free design, we do not assume any prior knowledge about transition probability.

\end{enumerate}

\subsection{MA2C for AVs}
In this paper, we extend the actor-critic network \cite{mnih2016asynchronous} to the multi-agent setting as a multi-agent actor-critic network (i.e., MA2C). MA2C improves the stability and scalability of the learning process by allowing certain communication among agents \cite{chu2020multi}. To take the advantage of homogeneous agents in the considered MARL setting, we assume all the agents share the same network structure and parameters, while they are still able to make different maneuvers according to different input states.
The goal in cooperative MARL setting is to maximize the global reward of all the agents. To overcome the communication overhead and the credit assignment problem \cite{sutton2018reinforcement}, we adopt the local reward design \cite{chen2021deep} as follows:
\begin{equation}\label{eqn:local_reward}
        r_{i, t} = \frac{1}{\mid{\nu_i}\mid} \sum_{j\in\nu_i} r_{j,t},
\end{equation}
where $\mid \nu_i \mid$ denotes the cardinality of a set containing the ego vehicle and its close neighbors. Compared to the global reward design previously used in \cite{kaushik2018parameter, dong2020drl}, the designed local reward design mitigates the impact of remote agents.

\begin{figure}[!ht]
    \centering
    \includegraphics[width=0.92\textwidth]{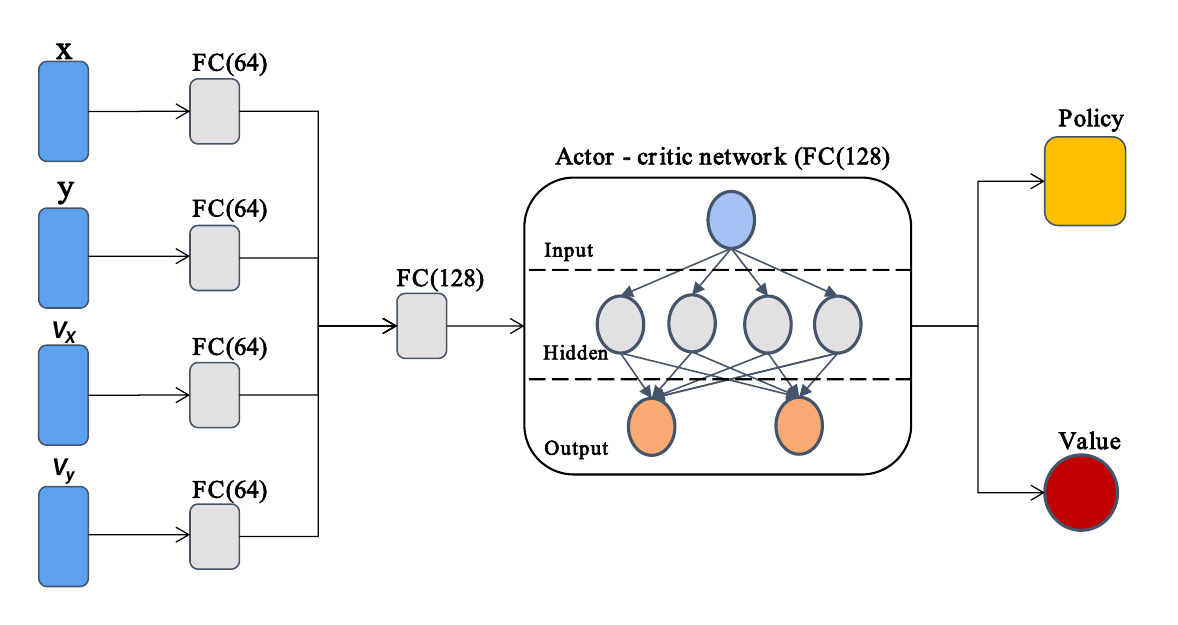}
    \caption{The architecture of the proposed MA2C network with shared actor-critic network design, where $x$ and $y$ are the longitudinal and lateral position of the observed vehicle relative to the ego vehicle, and $v_x$ and $v_y$ are the longitudinal and lateral speed of the observed vehicle relative to the ego vehicle.}
    \label{Backbone of MA2C Network}
\end{figure}

The backbone of the proposed MA2C network is shown in Fig.~\ref{Backbone of MA2C Network}, in which states separated by physical units are first processed by separate 64-neuron fully connected (FC) layers. Then all hidden units are combined and fed into the 128-neuron FC layer. Then the shared actor-critic network will update the policy and value networks with the extracted features. As mentioned in the recent literature \cite{chen2021deep}, the adopted parameter sharing scheme \cite{lin2018efficient} between the actor and value networks can greatly improve the learning efficiency.

The pseudo-code of the proposed MA2C algorithm is shown in Algorithm \ref{algo:marl_algo}. The hyperparameters include the (time)-discount factor $\gamma$, the learning rate $\eta$, the politeness coefficient $p$ and the epoch length $T$. Specifically, the agent receives the observation $O_{i,t}$ from the environment and updates the action by its policy (Line 3-6). After each episode is completed, the network parameters are updated accordingly (Line 9-11). If an episode is completed or a collision occurs, the ``DONE" signal is released and the environment will be reset to its initial state to start a new epoch (Line 13-14).

\begin{algorithm}[htb]  
  \caption{MARL for AVs.}  
  \label{algo:marl_algo}  
  \begin{algorithmic}[1]  
    \Require  $\gamma, \eta, p, T$.
    \Ensure   $\theta$. 
    \State {\bf Initialize}  $o_0, t \leftarrow 0$.
    
    \Repeat
       \For{$i \in V$}
       \State Observe $o_{i,t}$;
       \State Update $a_{i,t} \sim \pi_{\theta_{i,t}}$;
       \EndFor
       
       \State Update $t=t+1$;
       \If{DONE}
           \For{$i \in V$}
           \State Update $\theta_i \leftarrow \theta_i + \eta \nabla_{\theta_i} {J(\theta_i)}$;
           \EndFor
           \EndIf  
       \If{$t=T$}
           \State Initialize $o_{0},t \leftarrow 0$;
        \EndIf  
    
    \Until{Stop condition is reached}
\end{algorithmic}  
\end{algorithm}

\section{Experiments and Discussion}
\label{section:NUMERICAL EXPERIMENTS AND DISCUSSION}
In this section, we evaluate the performance of the proposed MARL algorithm in terms of training efficiency, safety, and driving comfort in the considered highway lane changing scenario shown in Fig.~\ref{lane-changing scene}. 

\subsection{HDV Models}
In this experiment, we assume that the longitudinal control of HDVs follows the Intelligent Driver Model (IDM) \cite{Treiber}, which is a deterministic continuous-time model describing the dynamics of the position and speed of each vehicle. It takes into account the expected speed, distance between the vehicles and the behavior of the acceleration/deceleration process caused by the different driving habits. In addition, Minimize Overall Braking Induced By Lane Change model (MOBIL) ~\cite{Kesting} is adopted for the lateral control. It takes vehicle acceleration as the input variable of the model and can work well with most car-following models. The acceleration expression is defined as follows:

\begin{equation}
\label{equation:safe}
    \tilde{a}_{n} \ge -b_{safe},
\end{equation}
where $\tilde{a}_{n}$ is the acceleration of the new follower after the lane change, and $b_{safe}$ is the maximum braking imposed to the new follower. If the inequality in Eqn.~\ref{equation:safe} is satisfied, the ego vehicle is able to change lanes. The incentive condition is defined as:

\begin{equation}
\label{equation:incentive}
{\underbrace{\tilde{a}_{c}-a_{c}}_{\text{\rm ego~vehicle}}}+p\big({\underbrace{\tilde{a}_{n}-a_{n}}_{\text{\rm new~follower}}}+{\underbrace{\tilde{a}_{o}-a_{o}}_{\text{\rm old~follower}}}\big) \ge \Delta a_{th},
\end{equation}
where $a$ and $\tilde{a}$ are the acceleration of the ego vehicle before and after the lane change, respectively, $\Delta a_{th}$ is the threshold that determines whether to trigger the lane change or not, and $p$ is a politeness coefficient that controls how much effect we want to take into account for the followers, where $p=1$ represents the most considerate drivers whose decision on change lanes may give way to the following blocked vehicles whereas $p=0$ characterizes the most aggressive drivers where the HDV makes selfish lane-changing decisions by only considering their own speed gains and ignoring other vehicles. The performance evaluation of different $p$ values is discussed in Section~\ref{subsection:verification of Driving Comfort}.

\subsection{Experimental Settings}
The simulation environment is modified from the gym-based highway-env simulator~\cite{highway-env}. We set the highway road length to $520~m$, and the vehicles beyond the road are ignored. The vehicles are randomly spawned on the highway with different initial speeds $25-30 ~m/s$ ($56~mph-67~mph$). 
The vehicle control sampling frequency is set as the default value of $5~Hz$. The motions of HDVs follow the IDM and MOBIL model, where the maximum deceleration for safety purposes is limited by $b_{safe}=-9~m/s^{2}$, politeness factor $p$ is 0, and the lane-changing threshold $\Delta a_{th}$ is set as $0.1~m/s^{2}$. 
To evaluate the effectiveness of the proposed methods, three traffic density levels are employed, which correspond to low, middle, high levels of traffic congestion, respectively. The number of vehicles in different traffic modes is shown in Table~\ref{tab:TrafficDensityModes}. 

\begin{table}[!ht]
\centering
\caption{Traffic density modes.}
\label{tab:TrafficDensityModes}
\begin{tabular}{cccc}
\hline
Traffic density modes & AVs & HDVs &Explanation \\ \hline
1                & 1-3 & 1-3 & low level \\ 
2              & 2-4 & 2-4  & middle level \\ 
3                & 4-6 & 4-6 & high level\\ \hline
\end{tabular}
\end{table}

We train the MARL algorithms for 1 million steps (10,000 epochs) by applying two different random seeds and the same random seed is shared among agents. We evaluate each model 3 times every 200 training episodes. The parameters $\gamma$ and learning rate $\eta$ are set as $0.99$ and $5 \times 10^{-4}$, respectively. The weighting coefficients in the reward function are set as $\omega_{s}=200$, $\omega_{d}=4$, $\omega_{v}=1$ and $\omega_{c}=1$, respectively. These experiments are conducted on a macOS server with a 2.7 GHz Intel Core i5 processor and 8GB of memory. 

\subsection{Results \& Analysis}
\subsubsection{Local v.s. Global Reward Designs}
\label{subsection:Local vs Global Rewards}
\begin{figure*}[!ht]  
    \centering  
    \includegraphics[width=1\textwidth]{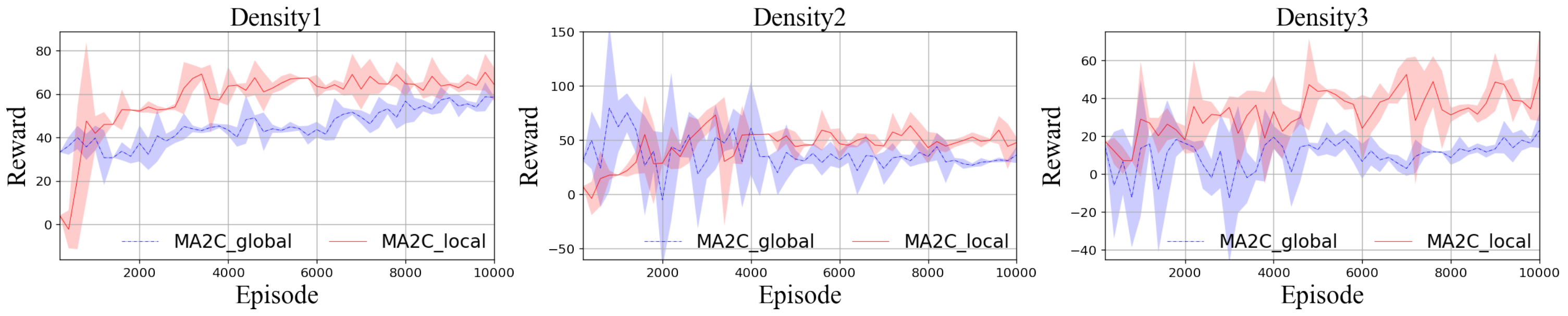}  
    \caption{Performance comparisons between local and global reward designs. The shaded region denotes the standard deviation over 2 random seeds.}  
    \label{Global vs Local}  
\end{figure*}

Fig.~\ref{Global vs Local} shows the performance comparison between the proposed local reward and the global reward design \cite{kaushik2018parameter, dong2020drl} (with shared actor-critic parameters). In all three traffic modes, the local reward design consistently outperforms the global reward design in terms of larger evaluation rewards and smaller variance. Although the global reward design outperforms the local reward design before 2000 epochs, the variance of the global reward is relatively large. In addition, the performance gaps are enlarged as the number of vehicles increases. This is due to the fact that the global reward design is more likely to cause credit assignment issues as mentioned in the significant monograph \cite{sutton2018reinforcement}.

\subsubsection{Sharing v.s. Separate Actor-critic Network}
\label{subsection:Sharing vs No-sharing multi-agent A2C Network}
\begin{figure*}[!ht]
    \centering  
    \includegraphics[width=1\textwidth]{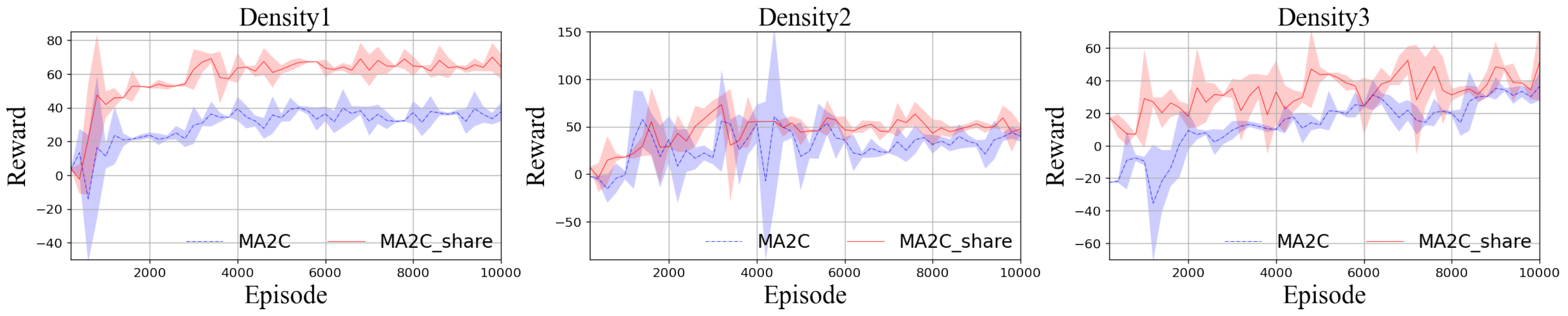}
    \caption{Performance comparisons between with and without actor-critic network sharing.} 
    \label{Sharing vs No-sharing}  
\end{figure*}

Fig.~\ref{Sharing vs No-sharing} shows the performance comparison between strategies with or without sharing the actor-critic network parameters during training. Obviously, sharing an actor-critic network has better performance than without sharing. Specifically, sharing actor-critic parameters in all three modes results in higher rewards and lower variance. The reason is that, in separate actor-critic networks, the critic network can only guide the actor network to the correct training direction until the critic network is well-trained which may take a long time to achieve. In contrast, the actor network can benefit from the shared state representation via the critic network in a shared actor-critic network \cite{chen2021deep, graesser2019foundations}.

\subsubsection{Verification of Driving Comfort}
\label{subsection:verification of Driving Comfort}
\begin{figure*}[!ht]
    \centering  
    \includegraphics[width=0.78\textwidth]{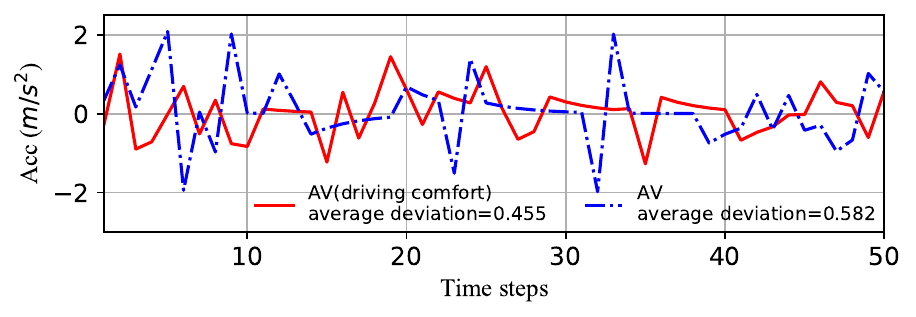}  
    \caption{Performance comparisons of acceleration between the reward design with or without comfort measurement.} 
    \label{acc}  
\end{figure*}

In this subsection, we evaluate the effectiveness of the proposed multi-objective reward function with the driving comfort in Eqn.~\ref{eqn:reward}. Fig.~\ref{acc} shows the acceleration and deceleration of the AV with or without the comfort measurement defined in Eqn.~\ref{eqn:comfort}. It is clear that the proposed reward design with the comfort measurement has a low variance (average deviation: $0.455 m/s^{2}$) and is more smooth than the reward design without comfort term (average deviation: $0.582m/s^{2}$), which shows the proposed reward design presents good driving comfort. 

\subsubsection{Adaptability of the Proposed Method}
\label{subsection:Adaptability of the proposed method}
\begin{figure*}[!ht]
    \centering  
    \includegraphics[width=1\textwidth]{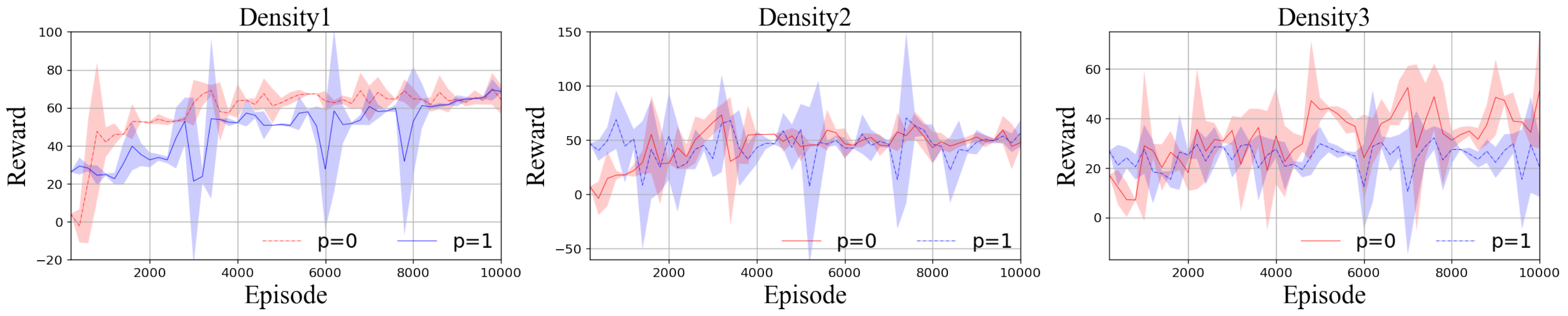}  
    \caption{Performance comparisons on different politeness coefficients $p$ under different traffic densities.} 
    \label{polite}  
\end{figure*}

In this subsection, we evaluate the proposed MA2C under different HDV behaviors, which is controlled by the politeness coefficient $p$ denoted in Eqn.~\ref{equation:incentive}, in which $p=0$ means the most aggressive behavior while $p=1$ represents the most polite behavior.
Fig.~\ref{polite} shows the training performance of two different HDV models (i.e., aggressive or politeness) under different traffic densities. It is clear that the proposed algorithm achieves scalable and stable performance whenever the HDVs take aggressive or courteous behaviors.

\subsubsection{Comparison with the state-of-the-art benchmarks}
\label{subsection:Comparison with Other Methods}
In order to demonstrate the performance of the proposed MARL approach, we compared it with several state-of-the-art MARL methods:
\begin{enumerate}
\item \textit{Multi-agent Deep Q-Network} (MADQN) \cite{Ji}: This is the multi-agent version of Deep Q-Network (DQN) \cite{mnih2013playing}, which is an off-policy RL method by applying a deep neural network to approximate the value function and an experience replay buffer to break the correlations between samples to stabilize the training.

\item \textit{Multi-agent actor-critic using Kronecker-Factored Trust Region} (MAACKTR): This is the multi-agent version of actor-critic using Kronecker-Factored Trust Region (ACKTR) \cite{Yuhuaiwu}, which is an on-policy RL algorithm by optimizing both the actor and the critic using Kronecker-factored approximate curvature (K-FAC) with trust region. 

\item \textit{Multi-agent Proximal Policy Optimization (MAPPO)} \cite{Schulman}: This is a multi-agent version of Proximal Policy Optimization (PPO) \cite{schulman2017proximal}, which improves the trust region policy optimization (TRPO) \cite{schulman2015trust} by using a clipped surrogate objective and adaptive KL penalty coefficient.

\item \textit{The Proposed MA2C}: This is our proposed method with the designed multi-objective reward function design, parameter sharing, and local reward design schemes.
\end{enumerate}

\begin{figure*}[!ht]  
    \centering  
    \includegraphics[width=1\linewidth]{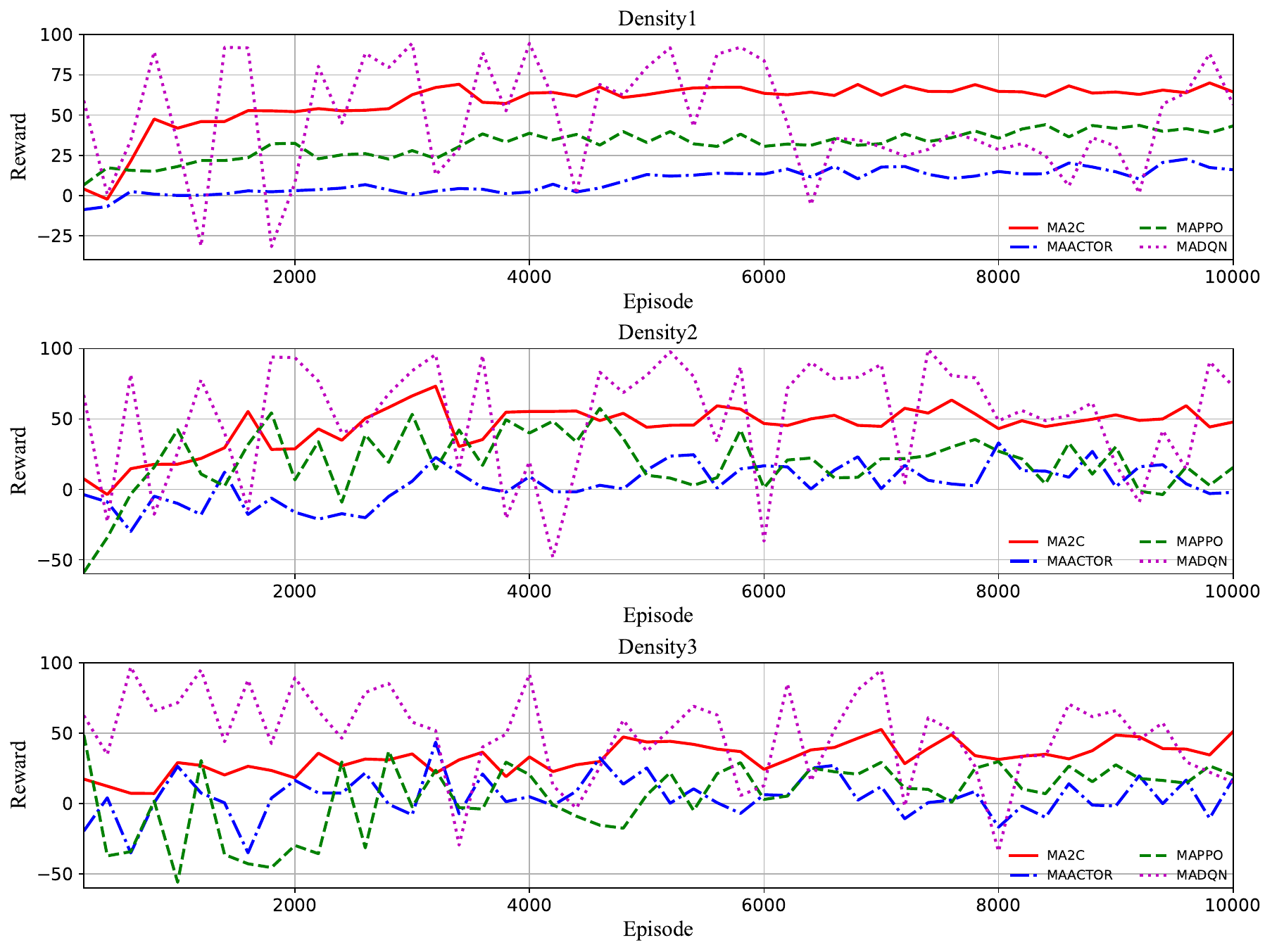}
    \caption{Performance comparisons on accumulated rewards in  MADQN, MA2C, MAACKTR, and MAPPO.}  
    \label{Comparison with Other Methods}  
\end{figure*}

Table~\ref{tab:MeanEpisodeReward} shows the average return for the MARL algorithms during the evaluation. Obviously, the proposed MA2C algorithm shows the best performance under the density1 scenario than other MARL algorithms. It also shows promising results on the density2 and density3 scenarios and outperforms MAACKTR and MAPPO algorithms. Note that even though MADQN shows a better average reward than the MA2C algorithm, it shows larger reward deviations which may cause unstable training and safety issues. MADQN depends on the current state during calculation, and cannot draw specific control strategies, and is not suitable for complex control with large state gaps. Indeed, MA2C appears more robust performances, which shows a very clear increasing and plateauing tendency. Similarly, the evaluation curves during the training process are shown in Fig.~\ref{Comparison with Other Methods}. As expected, the proposed MA2C algorithm outperforms other benchmarks in terms of evaluation reward and reward standard deviations.

\begin{table}[!ht]
\centering
\caption{Mean episode reward in different traffic flow scenario.}
\label{tab:MeanEpisodeReward}
\begin{tabular}{cccc}
\hline
Method & Density 1 & Density 2 &Density 3 \\ \hline
MADQN    &47.451 &51.568 &48.509 \\ 
&($\pm 27.948$)& ($\pm 32.943$) & ($\pm 24.078$) \\ 

MA2C & 58.000  & 44.744  & 32.579  \\
     & ($\pm 9.308$) & ($\pm 10.895$)  & ( $\pm 8.160$) \\
MAACKTR  &8.812 &3.759  &4.892 \\
 &($\pm 6.217$) &($\pm 10.858$)&($\pm 10.986$)\\

MAPPO    & 31.988 &19.300 &5.073 \\
&($\pm 6.567$) &($\pm 16.097$)	&($\pm 19.762$)\\ \hline
\end{tabular}
\end{table}

\subsubsection{Policy Interpretation}
\begin{figure}[!ht]
\centering

\subfigure[initial state]{
\begin{minipage}[t]{1\linewidth}
\centering
\includegraphics[width=1\linewidth]{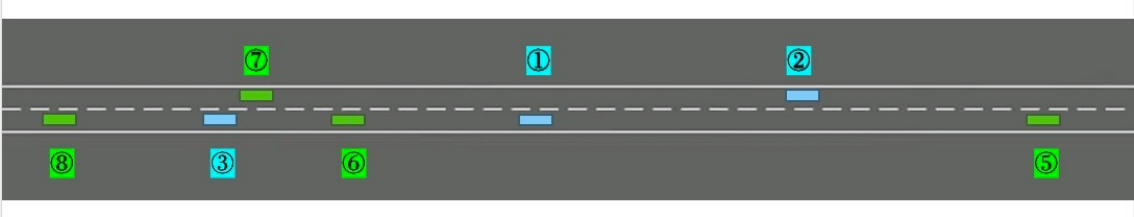}
\label{inital_state}
\end{minipage}%
}%
\quad

\subfigure[changing lanes]{
\begin{minipage}[t]{1\linewidth}
\centering
\includegraphics[width=1\linewidth]{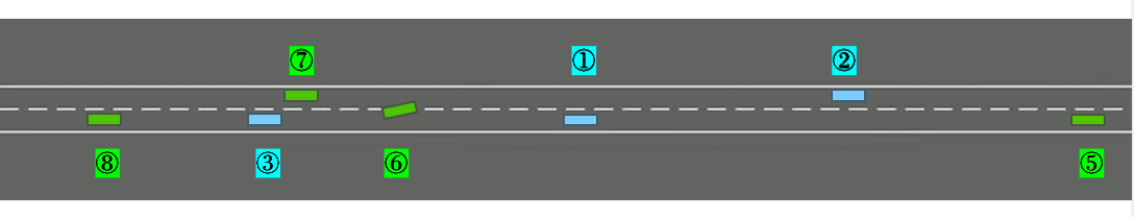}
\label{changing_lanes}
\end{minipage}%
}%
\quad

\subfigure[lane change completed]{
\begin{minipage}[t]{1\linewidth}
\centering
\includegraphics[width=1\linewidth]{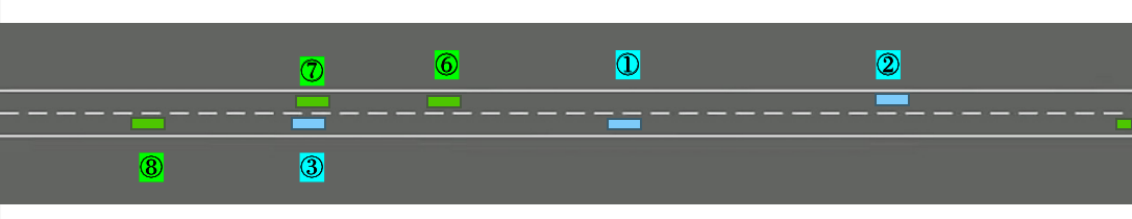}
\label{lane_change_completed}
\end{minipage}
}%

\centering
\caption{Lane change in simulation environment (vehicles \ding{192}-\ding{194}: HDVs, vehicles \ding{197}-\ding{199}: AVs).}
\label{Lane Change}  
\end{figure}

\begin{figure}[!ht]  
    \centering  
    \includegraphics[width=0.8\linewidth]{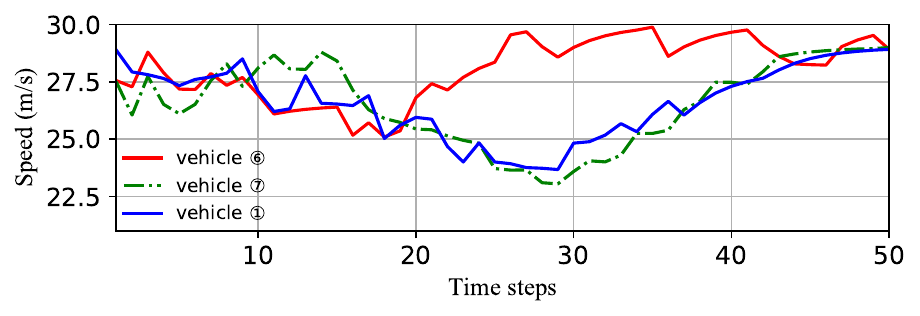}
    \caption{{Speeds of the AVs \ding{197}, \ding{198} and HDVs \ding{192}.}}  
    \label{Speeds of the AVs}  
\end{figure}

In this subsection, we attempt to interpret the learned AVs' behavior. Fig.~\ref{Lane Change} shows the snapshots during testing at time steps 20, 28, and 40. As shown in Fig.~\ref{inital_state}, ego vehicle \ding{197} attempts to make a lane change to achieve a higher speed. To make a safe lane change, ego vehicle \ding{197} and ego vehicle \ding{198} are expected to work cooperatively. Specially, the ego vehicle \ding{198} should slow down to make space for the ego vehicle \ding{197} to avoid collisions, which is also represented in Fig.~\ref{Speeds of the AVs}, where the ego vehicle \ding{198} starts to slow down at about 20-time steps. Then the ego vehicle \ding{197} begins to speed up to make the lane change as shown in Fig.~\ref{changing_lanes} and Fig.~\ref{Speeds of the AVs}. Meanwhile, the ego vehicle \ding{198} continues to slow down to ensure a safe headway distance with ego vehicle \ding{197} as shown in Fig.~\ref{Speeds of the AVs}. Fig.~\ref{lane_change_completed} shows the completed lane changes, at which time the ego vehicle \ding{198} starts to speed up. This demonstration shows the proposed MARL framework learns a reasonable and cooperative policy for ego vehicles.

\section{Conclusion}
\label{section:Conclusion}

In this paper, we formulated the highway lane-changing problem in mixed traffic as an on-policy MARL problem, and extended the A2C into the multi-agent setting, featuring a novel local reward design and parameter sharing schemes. Specifically, a multi-objective reward function is proposed to simultaneously promote the driving efficiency, comfort, and safety of autonomous driving. Comprehensive experimental results, conducted on three different traffic densities under different levels of HDV aggressiveness, show that our proposed MARL framework consistently outperforms several state-of-the-art benchmarks in terms of efficiency, safety, and driver comfort.

\bmhead{Acknowledgments}
The authors are grateful for the efforts of our colleagues in the Sino-German Center of Intelligent Systems, Tongji University. We are grateful for the suggestions on our manuscript from Dr. Qi Deng.

\bmhead{Authors' Contribution}
Wei Zhou, Dong Chen, Jun Yan, and Prof. Zhaojian Li participated in the framework design and manuscript writing, and Wei Zhou implemented experiments inspired by the encouragement and guidance of Dong Chen. Prof. Huilin Yin helped revise the manuscript. Prof. Wancheng Ge is the master supervisor of Wei Zhou in Tongji University which provides the opportunity to complete the research work.

\bmhead{Funding}
Jun Yan is supported by the National Natural Science
Foundation of China under Grant No. 61701348 hosted by Pro. Huilin Yin. Jun Yan, Prof. Huilin Yin, and Prof. Wancheng Ge are grateful for the generous support of T{\"{U}}V S{\"{U}}D. Wei Zhou is supported by the DAAD scholarship for a dual-degree program between Tongji University and Technical University of Munich.

\bmhead{Availability of data and materials}
Not applicable.

\bmhead{Code availability}
Not applicable.

\bmhead{Competing interests}
There are no conflicts of interest for this paper.

\bibliography{main}

\begin{thebibliography}{10}

\bibitem{Paden}
Brian Paden, Michal C{\'{a}}p, Sze~Zheng Yong, Dmitry~S. Yershov, and Emilio
  Frazzoli.
\newblock A survey of motion planning and control techniques for self-driving
  urban vehicles.
\newblock {\em {IEEE} Trans. Intell. Veh.}, 1(1):33--55, 2016.

\bibitem{desiraju2014minimizing}
Divya Desiraju, Thidapat Chantem, and Kevin Heaslip.
\newblock Minimizing the disruption of traffic flow of automated vehicles
  during lane changes.
\newblock {\em {IEEE} Trans. Intell. Transp. Syst.}, 16(3):1249--1258, 2015.

\bibitem{li2020cooperative}
Tingting Li, Jianping Wu, Ching-Yao Chan, Mingyu Liu, Chunli Zhu, Weixin Lu,
  and Kezhen Hu.
\newblock A cooperative lane change model for connected and automated vehicles.
\newblock {\em IEEE Access}, 8:54940--54951, 2020.

\bibitem{chen2020autonomous}
Dong Chen, Longsheng Jiang, Yue Wang, and Zhaojian Li.
\newblock Autonomous driving using safe reinforcement learning by incorporating
  a regret-based human lane-changing decision model.
\newblock In {\em American Control Conference (ACC)}, pages 4355--4361, 2020.

\bibitem{wang2019continuous}
Pin Wang, Hanhan Li, and Ching-Yao Chan.
\newblock Continuous control for automated lane change behavior based on deep
  deterministic policy gradient algorithm.
\newblock In {\em IEEE Intelligent Vehicles Symposium (IV)}, pages 1454--1460,
  2019.

\bibitem{xi2020efficient}
Chenyang Xi, Tianyu Shi, Yuankai Wu, and Lijun Sun.
\newblock Efficient motion planning for automated lane change based on
  imitation learning and mixed-integer optimization.
\newblock In {\em 23rd International Conference on Intelligent Transportation
  Systems (ITSC)}, pages 1--6, 2020.

\bibitem{wang2021harmonious}
Guan Wang, Jianming Hu, Zhiheng Li, and Li~Li.
\newblock Harmonious lane changing via deep reinforcement learning.
\newblock {\em {IEEE} Trans. Intell. Transp. Syst.}, 2021.

\bibitem{du2020cooperative}
Runjia Du, Sikai Chen, Yujie Li, Jiqian Dong, Paul Young~Joun Ha, and Samuel
  Labi.
\newblock A cooperative control framework for {CAV} lane change in a mixed
  traffic environment.
\newblock {\em CoRR}, abs/2010.05439, 2020.

\bibitem{Hoel}
Carl{-}Johan Hoel, Krister Wolff, and Leo Laine.
\newblock Automated speed and lane change decision making using deep
  reinforcement learning.
\newblock In Wei{-}Bin Zhang, Alexandre~M. Bayen, Javier J.~S{\'{a}}nchez
  Medina, and Matthew~J. Barth, editors, {\em 21st International Conference on
  Intelligent Transportation Systems (ITSC)}, pages 2148--2155, 2018.

\bibitem{vinyals2019grandmaster}
Oriol Vinyals, Igor Babuschkin, Wojciech~M Czarnecki, Micha{\"e}l Mathieu,
  Andrew Dudzik, Junyoung Chung, David~H Choi, Richard Powell, Timo Ewalds,
  Petko Georgiev, et~al.
\newblock Grandmaster level in starcraft ii using multi-agent reinforcement
  learning.
\newblock {\em Nat.}, 575(7782):350--354, 2019.

\bibitem{Chu}
Tianshu Chu, Jie Wang, Lara Codec{\`{a}}, and Zhaojian Li.
\newblock Multi-agent deep reinforcement learning for large-scale traffic
  signal control.
\newblock {\em {IEEE} Trans. Intell. Transp. Syst.}, 21(3):1086--1095, 2020.

\bibitem{lin2018efficient}
Kaixiang Lin, Renyu Zhao, Zhe Xu, and Jiayu Zhou.
\newblock Efficient large-scale fleet management via multi-agent deep
  reinforcement learning.
\newblock In {\em Proceedings of the 24th ACM SIGKDD International Conference
  on Knowledge Discovery \& Data Mining}, pages 1774--1783, 2018.

\bibitem{shalev2016safe}
Shai Shalev{-}Shwartz, Shaked Shammah, and Amnon Shashua.
\newblock Safe, multi-agent, reinforcement learning for autonomous driving.
\newblock {\em CoRR}, abs/1610.03295, 2016.

\bibitem{wangmulti}
Jiawei Wang, Tianyu Shi, Yuankai Wu, Luis Miranda-Moreno, and Lijun Sun.
\newblock Multi-agent graph reinforcement learning for connected automated
  driving.
\newblock In {\em Proceedings of the 37th International Conference on Machine
  Learning (ICML)}, 2020.

\bibitem{ha2020leveraging}
Paul Young~Joun Ha, Sikai Chen, Jiqian Dong, Runjia Du, Yujie Li, and Samuel
  Labi.
\newblock Leveraging the capabilities of connected and autonomous vehicles and
  multi-agent reinforcement learning to mitigate highway bottleneck congestion.
\newblock {\em CoRR}, abs/2010.05436, 2020.

\bibitem{palanisamy2020multi}
Praveen Palanisamy.
\newblock Multi-agent connected autonomous driving using deep reinforcement
  learning.
\newblock In {\em International Joint Conference on Neural Networks (IJCNN)},
  pages 1--7, 2020.

\bibitem{chen2021graph}
Sikai Chen, Jiqian Dong, Paul Ha, Yujie Li, and Samuel Labi.
\newblock Graph neural network and reinforcement learning for multi-agent
  cooperative control of connected autonomous vehicles.
\newblock {\em Comput. Aided Civ. Infrastructure Eng.}, 36(7):838--857, 2021.

\bibitem{ho2009lane}
Man~Lung Ho, Ping~T Chan, and AB~Rad.
\newblock Lane change algorithm for autonomous vehicles via virtual curvature
  method.
\newblock {\em Journal of advanced Transportation}, 43(1):47--70, 2009.

\bibitem{nilsson2016if}
Julia Nilsson, Jonatan Silvlin, Mattias Brannstrom, Erik Coelingh, and Jonas
  Fredriksson.
\newblock If, when, and how to perform lane change maneuvers on highways.
\newblock {\em {IEEE} Intell. Transp. Syst. Mag.}, 8(4):68--78, 2016.

\bibitem{nilsson2016lane}
Julia Nilsson, Mattias Br{\"a}nnstr{\"o}m, Erik Coelingh, and Jonas
  Fredriksson.
\newblock Lane change maneuvers for automated vehicles.
\newblock {\em {IEEE} Intell. Transp. Syst. Mag.}, 18(5):1087--1096, 2016.

\bibitem{chen2019attention}
Yilun Chen, Chiyu Dong, Praveen Palanisamy, Priyantha Mudalige, Katharina
  Muelling, and John~M Dolan.
\newblock Attention-based hierarchical deep reinforcement learning for lane
  change behaviors in autonomous driving.
\newblock In {\em Proceedings of the IEEE/CVF Conference on Computer Vision and
  Pattern Recognition (CVPR) Workshops}, pages 137--145, 2019.

\bibitem{dong2020drl}
Jiqian Dong, Sikai Chen, Paul Young~Joun Ha, Yujie Li, and Samuel Labi.
\newblock A drl-based multiagent cooperative control framework for {CAV}
  networks: a graphic convolution {Q} network.
\newblock {\em CoRR}, abs/2010.05437, 2020.

\bibitem{chen2021deep}
Dong Chen, Zhaojian Li, Yongqiang Wang, Longsheng Jiang, and Yue Wang.
\newblock Deep multi-agent reinforcement learning for highway on-ramp merging
  in mixed traffic.
\newblock {\em CoRR}, abs/2105.05701, 2021.

\bibitem{kipf2016semi}
Thomas~N. Kipf and Max Welling.
\newblock Semi-supervised classification with graph convolutional networks.
\newblock In {\em 5th International Conference on Learning Representations
  (ICLR)}, 2017.

\bibitem{mnih2013playing}
Volodymyr Mnih, Koray Kavukcuoglu, David Silver, Alex Graves, Ioannis
  Antonoglou, Daan Wierstra, and Martin~A. Riedmiller.
\newblock Playing atari with deep reinforcement learning.
\newblock {\em CoRR}, abs/1312.5602, 2013.

\bibitem{spaan2012partially}
Matthijs~TJ Spaan.
\newblock Partially observable markov decision processes.
\newblock In {\em Reinforcement Learning}, pages 387--414. Springer Verlag,
  2012.

\bibitem{mnih2016asynchronous}
Volodymyr Mnih, Adria~Puigdomenech Badia, Mehdi Mirza, Alex Graves, Timothy
  Lillicrap, Tim Harley, David Silver, and Koray Kavukcuoglu.
\newblock Asynchronous methods for deep reinforcement learning.
\newblock In {\em International conference on machine learning (ICML)}, pages
  1928--1937, 2016.

\bibitem{DBLP:journals/corr/abs-1811-07214}
Meha Kaushik, Nirvan Singhania, and K~Madhava Krishna.
\newblock Parameter sharing reinforcement learning architecture for multi agent
  driving behaviors.
\newblock {\em CoRR}, abs/1811.07214, 2018.

\bibitem{Schester2019LongitudinalPC}
Larry Schester and Luis~E. Ortiz.
\newblock Longitudinal position control for highway on-ramp merging: {A}
  multi-agent approach to automated driving.
\newblock In {\em 22nd {IEEE} Intelligent Transportation Systems Conference
  (ITSC)}, pages 3461--3468, 2019.

\bibitem{Palanisamy2020MultiAgentCA}
Praveen Palanisamy.
\newblock Multi-agent connected autonomous driving using deep reinforcement
  learning.
\newblock In {\em International Joint Conference on Neural Networks (IJCNN)},
  pages 1--7, 2020.

\bibitem{Wang2019ContinuousCF}
Pin Wang, Hanhan Li, and Ching{-}Yao Chan.
\newblock Continuous control for automated lane change behavior based on deep
  deterministic policy gradient algorithm.
\newblock In {\em {IEEE} Intelligent Vehicles Symposium (IV)}, pages
  1454--1460, 2019.

\bibitem{Mavrogiannis2020BGAPBA}
Angelos Mavrogiannis, Rohan Chandra, and Dinesh Manocha.
\newblock {B-GAP:} behavior-guided action prediction for autonomous navigation.
\newblock {\em CoRR}, abs/2011.03748, 2020.

\bibitem{chu2020multi}
Tianshu Chu, Sandeep Chinchali, and Sachin Katti.
\newblock Multi-agent reinforcement learning for networked system control.
\newblock In {\em 8th International Conference on Learning Representations
  (ICLR)}, 2020.

\bibitem{sutton2018reinforcement}
Richard~S Sutton and Andrew~G Barto.
\newblock {\em Reinforcement learning: An introduction}.
\newblock MIT press, 2018.

\bibitem{kaushik2018parameter}
Meha Kaushik, S.~Phaniteja, and K.~Madhava Krishna.
\newblock Parameter sharing reinforcement learning architecture for multi agent
  driving behaviors.
\newblock {\em CoRR}, abs/1811.07214, 2018.

\bibitem{Treiber}
M.~Treiber, A.~Hennecke, and D.~Helbing.
\newblock Congested traffic states in empirical observations and microscopic
  simulations.
\newblock {\em Physical Review E}, 62:1805--1824, 2000.

\bibitem{Kesting}
Arne Kesting, Martin Treiber, and Dirk Helbing.
\newblock Connectivity statistics of store-and-forward intervehicle
  communication.
\newblock {\em {IEEE} Trans. Intell. Transp. Syst.}, 11(1):172--181, 2010.

\bibitem{highway-env}
Edouard Leurent.
\newblock An environment for autonomous driving decision-making.
\newblock \url{https://github.com/eleurent/highway-env}, 2018.

\bibitem{graesser2019foundations}
Laura Graesser and Wah~Loon Keng.
\newblock {\em Foundations of deep reinforcement learning: theory and practice
  in Python}.
\newblock Addison-Wesley Professional, 2019.

\bibitem{Ji}
Guanglin Ji, Junyan Yan, Jingxin Du, Wanquan Yan, Jibiao Chen, Yongkang Lu,
  Juan Rojas, and Shing~Shin Cheng.
\newblock Towards safe control of continuum manipulator using shielded
  multiagent reinforcement learning.
\newblock {\em {IEEE} Robotics Autom. Lett.}, 6(4):7461--7468, 2021.

\bibitem{Yuhuaiwu}
Yuhuai Wu, Elman Mansimov, Roger~B. Grosse, Shun Liao, and Jimmy Ba.
\newblock Scalable trust-region method for deep reinforcement learning using
  kronecker-factored approximation.
\newblock In Isabelle Guyon, Ulrike von Luxburg, Samy Bengio, Hanna~M. Wallach,
  Rob Fergus, S.~V.~N. Vishwanathan, and Roman Garnett, editors, {\em Advances
  in Neural Information Processing Systems 30: Annual Conference on Neural
  Information Processing Systems (NeurIPS)}, pages 5279--5288, 2017.

\bibitem{Schulman}
John Schulman, Filip Wolski, Prafulla Dhariwal, Alec Radford, and Oleg Klimov.
\newblock Proximal policy optimization algorithms.
\newblock {\em CoRR}, abs/1707.06347, 2017.

\bibitem{schulman2017proximal}
John Schulman, Filip Wolski, Prafulla Dhariwal, Alec Radford, and Oleg Klimov.
\newblock Proximal policy optimization algorithms.
\newblock {\em CoRR}, abs/1707.06347, 2017.

\bibitem{schulman2015trust}
John Schulman, Sergey Levine, Pieter Abbeel, Michael Jordan, and Philipp
  Moritz.
\newblock Trust region policy optimization.
\newblock In {\em International conference on machine learning (ICML)}, pages
  1889--1897, 2015.

\end{thebibliography}

\end{document}